\documentclass{article}

    \PassOptionsToPackage{numbers, compress}{natbib}

    \usepackage[preprint]{neurips_2025}

\usepackage[utf8]{inputenc} 
\usepackage[T1]{fontenc}    
\usepackage[dvipsnames,table,xcdraw]{xcolor}
\usepackage[pagebackref,breaklinks,colorlinks,linkcolor=BrickRed,citecolor=RoyalBlue]{hyperref}
\usepackage{etoc}
\usepackage{url}            
\usepackage{booktabs}       
\usepackage{amsfonts}       
\usepackage{nicefrac}       
\usepackage{microtype}      
\usepackage{xcolor}         

\usepackage[table]{xcolor}
\usepackage{graphicx}
\usepackage{caption}
\usepackage{arydshln}
\usepackage{tikz}
\usepackage{wrapfig}
\usepackage{float}
\usepackage{multirow}
\usepackage{pifont}
\usepackage{makecell}
\newcommand{\cmark}{\ding{51}}%

\usepackage{amssymb} 
\usepackage{bbm}
\usepackage{algorithm}
\usepackage{algorithmicx}
\usepackage{amsmath}
\usepackage{algpseudocode}
\usepackage{subcaption}

\newcommand{\figref}[1]{Fig.~\ref{#1}}
\newcommand{\tabref}[1]{Tab.~\ref{#1}}
\newcommand{\eqnref}[1]{Eqn.~(\ref{#1})}
\newcommand{\agref}[1]{Algorithm.~(\ref{#1})}

\newcommand{\secref}[1]{Sec.~\ref{#1}}
\newcommand{\myPara}[1]{\noindent\textbf{#1}}

\def\MyMthd{LLaVA-Scissor}

\newcommand*\samethanks[1][\value{footnote}]{\footnotemark[#1]}

\newcommand{\tablestyle}[2]{\setlength{\tabcolsep}{#1}\renewcommand{\arraystretch}{#2}\centering\small}

\title{LLaVA-Scissor: Token Compression with Semantic Connected Components for Video LLMs  }

\author{Boyuan Sun$^{1,2}$\thanks{Equal contribution.} \quad Jiaxing Zhao$^{{2}}$\samethanks \quad Xihan Wei$^2$ \quad Qibin Hou$^1$\thanks{Corresponding author.}  \\
$^1$VCIP, School of Computer Science, Nankai University \\
$^2$Tongyi Lab, Alibaba Group  \\
 \textit{\normalsize {boyuansun}@mail.nankai.edu.cn, houqb@nankai.edu.cn}  \\
 {\textit{\normalsize{\{zjx244036, xihan.wxh\}@alibaba-inc.com}}}
 }

\begin{document}

\maketitle

\begin{abstract}
 In this paper, we present \MyMthd{}, a training-free token compression strategy designed for video multimodal large language models. Previous methods mostly attempt to compress tokens based on attention scores, but fail to effectively capture all semantic regions and often lead to token redundancy. Differently, we propose to leverage the Semantic Connected Components (SCC) approach that assigns
tokens to distinct semantic regions
 within the token set, ensuring comprehensive semantic coverage. The outcome is a two-step spatio-temporal token compression strategy that utilizes SCC in both spatial and temporal domains. This strategy can effectively compress tokens by representing the entire video with a set of non-overlapping semantic tokens. 
 We conduct extensive evaluations of the token compression capabilities of \MyMthd{} across diverse video understanding benchmarks, including video question answering, long video understanding, and comprehensive multi-choices benchmarks.
 Experimental results show that the proposed \MyMthd{} outperforms other token compression methods, achieving superior performance in various video understanding benchmarks, particularly at low token retention ratios. 
 
 Project page: \url{https://github.com/HumanMLLM/LLaVA-Scissor}.
\end{abstract}

\section{Introduction}
Recently, Video Large Language Models (VLLMs)~\cite{li2023videochat,maaz2023video,lin2023video,wang2024internvideo2,liu2024world, zhao2025facial, achiam2023gpt, zhao2025humanomni} have achieved remarkable progress due to the rapid advancement of Multimodal Large Language Models (MLLMs)~\cite{Qwen2-VL, li2024llava, chen2023internvl, gpt4v, zhu2023minigpt,chen2024expanding,ye2023mplugowl,bai2023qwen,dong2024internlm}.
Unlike processing a single image, VLLMs often require independent encoding of each video frame in a serialized sequence. As a result, even when only a small number of frames are sampled from a video, a large number of visual tokens are generated. 
Although some methods attempt to reduce the number of visual tokens from the perspective of network architecture~\cite{li2024llamavid, xue2024xgenmmblip3familyopen, ye2024voco} or trainable modules~\cite{blip3video-xgenmmvid, zhang2023vision, allakhverdov2025enoughadaptivetokenreduction}, these approaches not only involve additional training costs but also face limitations in portability due to their specialized structures. Therefore, training-free token reduction strategies during the inference phase are essential for more efficient and scalable video processing.

\begin{figure}[t]
    \centering
    \includegraphics[width=0.95\linewidth]{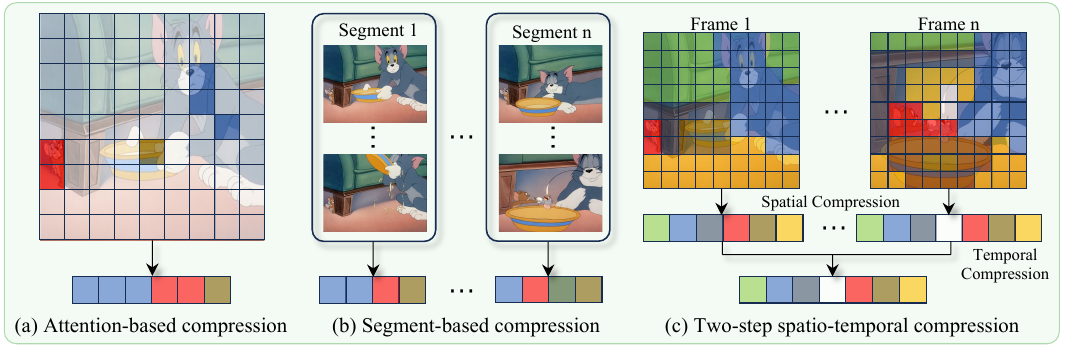}
    \vspace{-0.1cm}
    \caption{\textbf{Illustration  of different token compression paradigms.} $\square$ denotes video tokens, with color representing different semantics. (a) Attention-based methods fail to cover all semantic regions. 
    (b) Segment-based methods introduce temporal redundancy by stacking tokens from each segment. (c) Our two-step spatio-temporal compression strategy is able to identify unique semantic information within each frame and eliminate temporal redundancy, resulting in non-overlapping video tokens.}
    \label{fig: pattern} 
\end{figure}

Selecting the most representative tokens~\cite{fayyaz2022ats, shen2024tempme, bolya2022tome} from all tokens is one of the most common paradigms for token reduction. Previous methods designed for image token compression~\cite{yang2024visionzip, shang2024LLaVAPruMerge} often focus on leveraging attention scores to identify the most important tokens. However, as shown in \figref{fig: pattern}(a), attention-based approaches~\cite{chen2024image, zhang2024sparsevlm, xing2024pyramiddrop, liu2024multi} tend to prioritize only key objects, which can result in an incomplete representation of all semantics and the repeated selection of key semantics at the same time.
Therefore, to identify all distinct semantic regions that can effectively represent the entire video with less redundancy, we propose the Semantic Connected Components (SCC) strategy. By measuring pairwise similarity between tokens, SCC can partition the tokens into non-overlapping regions throughout identifying connected components, and 
use tokens to represent distinct semantic regions. 
Notably, SCC does not require the tokens to be positional adjacent, allowing it to capture global semantic relations throughout the entire token sequence, regardless of spatial positioning. This enables a more comprehensive and efficient representation of video content.

Considering that image-based methods fail to account for the temporal redundancy across video frames, some approaches~\cite{tao2024dycoke, huang2024prunevid, liu2025hybrid} begin to design compression strategies specifically for videos. 
As shown in \figref{fig: pattern}(b), these methods often focus on segmenting the video and applying inter-segment token compression~\cite{shen2025fastvid, guo2025fila, videollamb}, or compressing tokens based on fixed pixel positions across time~\cite{shen2024longvu}. However, they overlook the fact that semantically similar regions may not be temporally connected or maintain spatial consistency over time, which introduces potential redundancy.
To tackle this, we introduce \MyMthd{}, a two-step spatio-temporal token compression strategy as shown in \figref{fig: pattern}(c) with the help of SCC. Specifically, we first identify all unique semantic regions within the spatial domain of each video frame. Then, we assess the SCC again to remove temporal redundancy over semantic regions across frames and perform a further fusion. The result is a set of tokens that can effectively represent the entire video, without redundancy, and with each token encapsulating distinct semantic information from both spatial and temporal perspectives.

We evaluate our \MyMthd{} on three video question answering benchmarks, four long video understanding benchmarks, and the multi-choice benchmark MVBench leveraging an enhanced LLaVA-OneVision model as the base model. The results demonstrate that our method consistently outperforms other token compression approaches, particularly at lower token retention ratios.
Finally, we analyze the reducing law in video token compression in \secref{sec: reducing law} by evaluating performance changes across benchmarks as the token retention ratio decreases. The results demonstrate that the redundancy of video tokens indeed exists, and our method, compared to others, is more effective in preserving key semantics at lower retention ratios, leading to superior performance.

Our contributions can be summarized as follows:
\begin{itemize}
    \item We point out that existing attention score-based methods
    fail to fully represent the entire token set and propose Semantic Connected Components (SCC), a token compression strategy that captures all distinct semantic regions within the token set.
    \item We propose \MyMthd{}, a two-step spatio-temporal token compression designed for video MLLMs, which can generate a more comprehensive and efficient representation of video content. Experiments show that \MyMthd{} outperforms other token compression methods on various video understanding benchmarks.
\end{itemize}

\section{Related Work}
\subsection{Video Large Language Models}
Benefiting from the rapid advancement of Large Language Models (LLMs)~\cite{brown2020language,ouyang2022training,chatgpt,vicuna2023,touvron2023llama2,jiang2024mixtral,guo2025deepseek,liu2024deepseek,qwen2.5}, a wide range of powerful proprietary MLLMs~\cite{gpt4v, openai2023gpt4, Claude2024, geminiteam2024geminifamilyhighlycapable, openai2024gpt4o} and open-source community MLLMs~\cite{liu2024visual, liu2023llava, liu2023llava1.5, zhang2024llavanext-video,li2024llavanext-ablations, ye2023mplug, ye2023mplugowl2, gao2023llama} have emerged. 
Among them, Video Large Language Models (VLLMs)~\cite{zhao2025llava, zhao2025r1, zhang2024llavanextvideo} tailored for video understanding have gained increasing attention.
Typical VLLMs~\cite{lin2023video, damonlpsg2023videollama, Maaz2023VideoChatGPT} encode video frame sequences into raw video tokens with visual encoder and projector, and then 
feed them into LLMs along with user instructions to generate responses.
Methods such as MovieChat~\cite{song2023moviechat}, TimeChat~\cite{Ren2023TimeChat} and TimeSuite~\cite{ zeng2024timesuite} develop memory modules and timestamp-aware encoders 
for capturing better context.
LLaVA-OneVision~\cite{li2024llavaone} proposes a unified model capable of handling images, videos, audios, and other modalities simultaneously.
LLaVA-Next-Video~\cite{zhang2024videoinstructiontuningsynthetic} significantly improves model performance by leveraging large-scale synthetic data.
However, since the sequential encoding of frames leads to an increase in token numbers, existing VLLMs struggle in processing long video and computational efficiency, emphasizing the necessity of token reduction.

\subsection{Token Reduction in MLLMs}
Token reduction~\cite{bolya2022tome, rao2021dynamicvit, koner2024lookupvit, liang2022not} is essential for efficient inference and longer visual sequences, and has been extended to the MLLMs domain~\cite{ye2024mplugowl3longimagesequenceunderstanding, shi2023crossget, han2025filtercorrelatecompresstrainingfree, chen2024efficient, TokenPacker}. Approaches such as LLaMA-VID~\cite{li2024llamavid} and LongVA~\cite{zhang2024longcontexttransferlanguage} design token-efficient architectures, while other methods like LongVU~\cite{shen2024longvu}, VideoLLaMA~\cite{damonlpsg2024videollama2, damonlpsg2025videollama3}, and VideoLLaMB~\cite{videollamb}  reduce token numbers through projector-level modifications. 
Pooling-based approaches~\cite{xu2024pllava, yao2024deco, weng2024longvlmefficientlongvideo} are also widely used.
Additionally, some methods~\cite{wang2024lifelongmemoryleveragingllmsanswering, zhang2023simple, wang2024videoagentlongformvideounderstanding} explore agent-based techniques to convert videos into textual descriptions.

As for training-free strategies, the most common approach is to select the important tokens from the token set~\cite{shang2024LLaVAPruMerge, yang2025topv, wang2025dymudynamicmergingvirtual, Ye2024ATPLLaVAAT}. 
Some methods leverage the attention score of [CLS] token~\cite{yang2024visionzip, zhang2024fastervlm, zhang2024token, wang2024cls} to measure the importance of each token.
FastV~\cite{chen2024image} proposes a strategy to select key tokens during the prefilling stage based on attention maps, while VTW~\cite{lin2024boosting} introduces an aggressive approach that directly removes visual tokens after a certain decoder layer.
However, these image token compression methods overlook the temporal redundancy across video frames. 
DyCoke~\cite{tao2024dycoke} uniformly divides videos into segments and performs intra-segment compression by fixed even-odd token grouping. PruneVID~\cite{huang2024prunevid} and FastVID~\cite{shen2025fastvid} first segment the video based on scene boundaries and cluster video tokens within each segment. However, these segment-based methods overlook the fact that semantically similar information may not be temporally adjacent or spatially consistent, leading to redundancy when stacking tokens from each segment, as shown in \figref{fig: pattern}(b).
Different from them, \MyMthd{} performs token compression across both spatial and temporal dimensions, representing the entire video with a set of semantically non-overlapping video tokens.

\section{\MyMthd{}}

In this section, we first propose the Semantic Connected Components (SCC) approach to identify distinct semantic regions within a given token set by utilizing connected components. Then, we propose a two-step scheme that applies the SCC strategy in spatial and temporal domains to achieve effective token compression for video understanding.

\subsection{Token Compression via Semantic Connected Components (SCC)}

\label{sec: Connected Components}
As shown in \figref{fig: pattern}(a), prior attention-based token selection approaches tend to select redundant semantic regions while overlooking others, making it challenging to obtain a representative and comprehensive token set.
Different from them, we tend to identify all unique semantic regions and retain one single token for each distinct region. 

Given a set of tokens $ \mathbf{K} = \{\mathbf{k}_1, ..., \mathbf{k}_{\mathbf{N}}\}\in \mathbb{R}^{\mathbf{N}\times d}$, we first compute the pairwise similarity between each token and then convert it into a binary map $\mathcal{A}$ based on a threshold $\tau$, as shown in \eqnref{eqn: adjacency matrix}:
\begin{equation} 
\label{eqn: adjacency matrix}
    \mathcal{A} = (\frac{\mathbf{K}\cdot \mathbf{K}^T}{||\mathbf{K}||_{dim=1}\cdot||\mathbf{K}||_{dim=1}} > \tau)
    \in \mathbb{R}^{\mathbf{N}\times \mathbf{N}}.
\end{equation}

In the binary similarity map $\mathcal{A}$, $\mathcal{A}(i, j)$ indicates the similarity between any two tokens $ \mathbf{k}_i$ and $ \mathbf{k}_j$. Therefore, if we consider each token as a vertex and treat $\mathcal{A}$ as an adjacency matrix representing the connectivity between vertices, we can construct a graph based on the similarity relationships inherent in $\mathcal{A}$. 
In this case, the problem of finding all unique semantic regions can be transformed into the task of identifying all connected components in the graph using the adjacency matrix $\mathcal{A}$.

To tackle this, we present an approximate method for computing the connected components.
Specifically, given the adjacency matrix $\mathcal{A}$, we first sample a subset of vertices from the $N$ vertices based on an error tolerance $\epsilon$. The number of vertices sampled is determined as follows:
\begin{equation}
\label{eqn: sample}
    \mathbf{N}' = \min(\mathbf{N}, \lceil \frac{\log(\mathbf{N})}{\epsilon^2}\rceil).
\end{equation}
For each sampled vertex, we 
identify all its neighbors based on the adjacency matrix $\mathcal{A}$ and 
employ the union-find data structure with path compression and union-by-rank
to identify all connected components in the graph. Note that when the connected components extracted from $\mathbf{N'}$ sampled vertex do not cover all $\mathbf{N}$ vertices, each uncovered vertex would be viewed as a separate connected component. Additionally, to preserve the relative positional relationships among tokens, we sort the connected components based on the vertex ID with the highest degree within each cluster. Detailed data structure and approximate connected components algorithm are detailed in \secref{sec: algorithms} of appendix.

The sorted connected components set is denoted as  $C = \{C_1, ..., C_M\}$, where $M$ indicates there are totally $M$ connected components in the graph, and
each $C_i$ in $ C $ consists of $|C_i|$ unique vertices.
As expressed in \eqnref{property}, since connected components can partition a graph with N vertices into M disjoint subgraphs, no two connected components intersect, and the total number of vertices across all connected components equals the number of vertices in the graph.
\begin{equation}
\label{property}
   \forall i,j \in [1, M], C_i \cap C_j = \varnothing ; \sum^{M}_{i=1}|C_i| = \mathbf{N}.
\end{equation}
This property suggests that the token set $\mathbf{K}$ can be partitioned into $M$ distinct semantic regions throughout connected components $C$, which can encompass all non-overlapping semantics indicated by similarity map $\mathcal{A}$.
Therefore, we can compress the token set $\mathbf{K}$ into $\mathbf{K}'$ by aggregating the tokens within each $C_i$ in $C$ 
to avoid redundancy, detailed in \eqnref{eqn: merge}:
\begin{equation}
\label{eqn: merge}
     \mathbf{k}'_i = \frac{1}{|C_i|}\sum^{|C_i|}_{j=1}{\mathbf{k}_{i_j}}, i = 1, ..., M; \mathbf{K'} = \text{concat}[\mathbf{k}'_1, ... ,\mathbf{k}'_M] .
\end{equation}
This process constructs a representative token for each unique semantic region, thus transforming the original token set $ \mathbf{K} \in \mathbb{R}^{\mathbf{N}\times c} $ into $ \mathbf{K}' \in \mathbb{R}^{M\times c} $. 
Leveraging this characteristic, we are able to get representative tokens from a token set, regardless of their positional adjacency. 

\begin{figure}[t]
    \vspace{-12mm}
    \centering
    \includegraphics[width=0.99\linewidth]{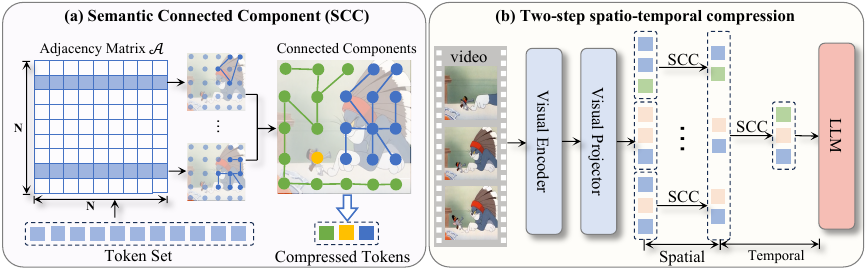}
    \vspace{-0.1cm}
    \caption{\textbf{Pipeline of \MyMthd{}.} (a) The Semantic Connected Components (SCC) compress tokens by extracting connected components from the token set. (b) The two-step spatio-temporal compression strategy that extracts unique semantics by leveraging SCC both spatially and temporally.}
    \label{fig: method} 
    \vspace{-10pt}
\end{figure}

\begin{table}[t]
\vspace{-5mm}
    \centering
    \small
    \setlength{\abovecaptionskip}{2pt}
    \definecolor{lightlightgray}{gray}{0.8}
    \tablestyle{2.3pt}{1.0}
    \begin{tabular}{lcccccccccccc}
        \toprule
        \multirow{2}{*}{Method} &  LMM &  Retention  & \multicolumn{2}{c}{ActivityNet} & \multicolumn{6}{c}{Video-ChatGPT} & Next- & \textbf{Avg.}\\
         \cmidrule(lr){4-5} \cmidrule(lr){6-11}
        & Size & Ratio  & Acc. & Score & CI & DO & CU & TU & CO & Avg. & QA &(\%) \\
        \midrule
        LLaVA-OneVision~\cite{li2024llavaone} & 7B & 100\% & 48.09 & 3.47 & 3.37 & 3.78 & 3.52 & 3.02 & 2.63 & 3.26 & 81.33 & 100\% \\
        \midrule
        FastV~\cite{chen2024image}  & 7B & 50\%
        & \textbf{47.95} & 3.47& 3.36 & \textbf{3.77} & 3.50 & 2.99 & 2.57 & 3.24 & \underline{81.11} & 99.4\%\\
        Dycoke~\cite{tao2024dycoke}  & 7B & 50\%
        & 47.88& 3.47 & 3.33 & 3.76 & \underline{3.51} & \underline{3.01} & 2.58 & 3.24 & 81.06 & 99.3\%\\ 
        PLLaVA~\cite{xu2024pllava} & 7B & 50\%
        & 47.59 & 3.45&\underline{ 3.36}& 3.73& \textbf{3.52}& 3.00& \textbf{2.66}& 3.25 & 81.04 & \underline{99.7\%}\\
        VisionZip~\cite{yang2024visionzip} & 7B & 50\%
        & 45.42& \underline{3.47} & 3.16 & 3.63 & 3.34 & 2.75 & 2.61 & 3.10 & 78.46 & 95.7\%\\  
        \rowcolor[HTML]{EFEFEF} \textbf{\MyMthd{}}  & 7B & 50\%
        &\underline{47.89} & \textbf{3.47 }& \textbf{3.37} & \underline{3.76} & 3.47 & \underline{3.00} & \underline{2.65} & 3.25 & \textbf{81.12 }& \textbf{99.7\%}\\
        \midrule
        FastV~\cite{chen2024image}  & 7B & 35\%
        & 47.83 & 3.46& 3.32 & 3.74 &\underline{ 3.47} & 2.97 & 2.61 & 3.22 & 80.49 & 99.0\%\\
        VisionZip~\cite{yang2024visionzip} & 7B & 35\%
        & 44.69 & 3.46 & 3.13 & 3.61 & 3.31 & 2.71 & 2.57 & 3.07 & 77.72 & 94.8\% \\
        Dycoke~\cite{tao2024dycoke}  & 7B & 35\%
        & 47.81& 3.45 & 3.31 & 3.74 & 3.46 & 2.98 & 2.54 & 3.21 & \textbf{80.86} & 98.6\%\\    
        PLLaVA~\cite{xu2024pllava} & 7B & 35\%
        & 47.23& 3.42 & 3.26 & 3.70 & 3.39 & 2.92 & 2.59 & 3.17 & 79.66 & 97.6\%\\  
        \rowcolor[HTML]{EFEFEF} \textbf{\MyMthd{}}  & 7B & 35\%
        &\textbf{47.88} & \textbf{3.47} &\underline{ 3.32} & \underline{3.75} & 3.46 & \textbf{2.99} & \textbf{2.63} & \textbf{3.23} & \underline{80.83} & \textbf{99.2\%}\\
        \rowcolor[HTML]{EFEFEF} \textbf{\MyMthd{}}  & 7B & 25\%
        &47.79 & \underline{3.47} & \textbf{3.33} & \textbf{3.76} & \textbf{3.47} & \underline{2.98} & \underline{2.62 }& \underline{3.23} & 80.66 & \underline{99.2\%}\\
        
        \rowcolor[HTML]{EFEFEF} \textbf{\MyMthd{}}  & 7B & 20\%
        &\underline{47.85} & 3.46 & 3.31 & 3.72 & 3.43 & 2.96 & 2.57 & 3.20 & 80.57 & 98.5\%\\
 \midrule       
        FastV~\cite{chen2024image}  & 7B & 10\%
        & 44.95 & 3.38& 3.04 & 3.60 & 3.28 & 2.80 & 2.49 & 3.04 & 78.76 & 94.2\%\\
        PLLaVA~\cite{xu2024pllava} & 7B & 10\%
        &45.28 & 3.37 & 3.11 & 3.56 & 3.25& 2.78& \textbf{2.55} & 3.05 & 77.87 & 94.4\% \\
        VisionZip~\cite{yang2024visionzip} & 7B & 10\%
        & 38.58& 3.30 & 2.65 & 3.09 & 2.73 & 2.31 & 2.42 & 2.64 & 65.09 & 82.7\%\\  
        \rowcolor[HTML]{EFEFEF} \textbf{\MyMthd{}}  & 7B & 10\%
        &\textbf{47.75} & \textbf{3.46} & \textbf{3.26} & \textbf{3.68} & \textbf{3.41} & \textbf{2.90} & 2.52 & \textbf{3.15} & \textbf{80.03} & \textbf{97.5\%}\\
        \rowcolor[HTML]{EFEFEF} \textbf{\MyMthd{}}  & 7B & 7\%
        &\underline{46.25} & \underline{3.42} & \underline{3.21} & \underline{3.63}&\underline{3.32} & 2.82 & 2.51 & \underline{3.10} & \underline{79.11} & \underline{95.8\%}\\
        \rowcolor[HTML]{EFEFEF} \textbf{\MyMthd{}}  & 7B & 5\%
        &46.09 & 3.41 & 3.16 & 3.60 & 3.30 & \underline{2.83}& \underline{2.53}& 3.08 & 78.82 & 95.5\%\\
        \bottomrule
    \end{tabular}
    \caption{Comparison of state-of-the-art token compression strategies under different token retention ratios on video question-answering benchmarks. 
    }
    \vspace{-15pt}
    \label{tab: VQA}
\end{table}

\subsection{Two-step Spatio-Temporal Token Compression}
\label{sec: two-step}
Equipped with the SCC strategy, we further propose a two-step token compression strategy across both spatial and temporal dimensions.
Given a video $\mathbf{V} = \{\mathbf{v}_1, ..., \mathbf{v}_n\}$ consisting of $n$ frames, the visual video tokens $\mathbf{T} = \{\mathbf{t}_1, ..., \mathbf{t}_n\} \in \mathbb{R}^{n\times m\times d}$ can be derived by processing it through the visual encoder and the visual projector, where $m$ denotes the number of tokens per frame and $d$ is the hidden state size. 
We first  determine the representative tokens for each frame in the spatial dimension. For each frame $ \mathbf{t}_i \in \mathbf{T} $ containing $ m'_i $ independent semantic regions, we use $ \mathbf{t}'_i $ to denote the representative tokens obtained through the SCC strategy. 
\begin{equation}
    \mathbf{t}'_i = \text{SCC} ( \mathbf{t}_i) \in \mathbb{R}^{m'_i \times d}, i= 1, ..., n.
\end{equation}
After spatial compression, we obtain the representative tokens for every frame in the video. Subsequently, we concatenate all $ t'_i $ together to further consider 
the redundancy temporally.
\begin{equation}
    \mathbf{T}' = \text{concat}[\mathbf{t}'_1, ..., \mathbf{t}'_n] \in \mathbb{R}^{M'\times d}; M' = \sum_{i=1}^{n}m'_i .
\end{equation}
In $ \mathbf{T}' $, all representative tokens from each frame of the video are included. However, distinct semantic regions within the spatial dimension of each frame may still share overlaps across frames. That is semantic regions appearing in one frame may also appear in other frames. Therefore, to avoid redundancy as much as possible, similar to spatial fusion, we apply the SCC strategy again temporally to further compress similar representative tokens across the sequence:
$\mathbf{T^r} = \text{SCC} ( \mathbf{T'}) \in \mathbb{R}^{M \times d}.$
The representative tokens $\mathbf{T^r}$ comprehensively cover all spatio-temporally distinct semantic regions without redundancy or overlap. Treating them as retained tokens, we select the most suitable semantic regions for all tokens inspired by ToMe~\cite{bolya2022tome}.

Specifically, we consider all tokens $ \mathbf{T^{a}} = \text{concat}[\mathbf{t}_1, ..., \mathbf{t}_n] \in \mathbb{R}^{(n\cdot m)\times d} $ as the source tokens, and the retained tokens $ \mathbf{T^{r}} $ as the target tokens. As shown in \eqnref{tome sim}, we first compute the similarity map and find the most similar token $ \mathbf{t}^{\mathbf{r}}_j $ in $ \mathbf{T^{r}} $ for each token $ \mathbf{t}^{\mathbf{a}}_i $ in $ \mathbf{T^{a}} $, storing the indices in $ I $.
\begin{equation}
\label{tome sim}
\begin{aligned}
     \mathcal{S} = \frac{ \mathbf{T^{a}}  \cdot (\mathbf{T^{r}})^{\top}}{||\mathbf{T^{a}}||\cdot ||\mathbf{T^{r}}||} \in \mathbb{R}^{(n\cdot m) \times M};
     I = \text{argmax}_{i\in M} \mathcal{S}(i) \in\mathbb{R}^{(n\cdot m)}.   
\end{aligned}
\end{equation}
$ I $ is a vector that stores the index of the most similar target token in $\mathbf{T^r}$ for each token in $ \mathbf{T^{a}} $. Specifically, $ I_i = j $ indicates that $ \mathbf{t}^\mathbf{a}_i $ is closest to $ \mathbf{t}^\mathbf{r}_j $. Finally, for each source token $ \mathbf{t}^\mathbf{a}_i \in \mathbf{T^{a}} $, we assign it to the most similar target token $ \mathbf{t}^\mathbf{r}_j \in \mathbf{T^r} $ according to $I$ and perform an average merge to obtain the final compressed token $ \mathbf{t}^{\mathbf{fin}}_j $.
\begin{equation}
    \mathbf{t}^{\mathbf{fin}}_j = \frac{ \sum^{n\cdot m}_{i=1} \mathbf{t}^\mathbf{a}_i \mathbbm{1}(I_i = j) + \mathbf{t}^{\mathbf{r}}_j }{\sum^{n\cdot m}_{i=1} \mathbbm{1}(I_i = j) + 1}, j = 1, ..., M.
\end{equation}
We obtain the final merged tokens $ \mathbf{T}^{\mathbf{fin}} = \text{concat}[\mathbf{t}^{\mathbf{fin}}_1, ..., \mathbf{t}^{\mathbf{fin}}_{M}] \in \mathbb{R}^{M\times d} $ to represent the entire video. 

\section{Experiments}
\subsection{Experiment Setup}
\label{sec: setup}
\myPara{Implement details.}
Similar to previous approaches, we choose LLaVA-OneVision~\cite{li2024llavaone} as the base model architecture, which originally utilizes the CLIP~\cite{radford2021clip} as the visual encoder and Qwen 2~\cite{qwen2} as the LLM. However, considering the rapid advancements in MLLM, its architecture and performance are no longer the most cutting-edge. For instance, SIGLIP (so400m-patch14-384)~\cite{zhai2023sigmoid} is regarded as a superior vision encoder in many methods~\cite{apollo}. Therefore, we replace the vision encoder in the original model with SIGLIP, employ the more advanced Qwen 2.5~\cite{qwen2.5} as the LLM, and retrain an enhanced version of the LLaVA-OneVision model using open-sourced Oryx~\cite{liu2024oryx} data.

Equipped with the enhanced LLaVA-OneVision model, we implement the proposed method and compare it with other approaches based on this model for a fair comparison. We primarily control the retention ratio through the similarity threshold $\tau$. Unless otherwise specified, we set $\epsilon$ to 0.05. All evaluation experiments in this paper are conducted on a single NVIDIA A100 GPU.

\myPara{Benchmarks and compared approaches.}
We validate the effectiveness of \MyMthd{} on a variety of benchmarks, including Video Question-Answering Benchmarks (ActivityNet-QA~\cite{yu2019activityqa}, Video-ChatGPT~\cite{Maaz2023VideoChatGPT}, Next-QA~\cite{xiao2021next}), Long Video Benchmarks (Egoschema~\cite{mangalam2023egoschema}, MLVU~\cite{MLVU}, Video-MME~\cite{fu2024videomme}, VideoMMMU~\cite{hu2025videommmu}), and the comprehensive multi-choice benchmark MVBench~\cite{li2024mvbench}. A detailed introduction to each dataset is provided in \secref{sec: benchmarks} of appendix.

We compare against several open-source representative methods: the image-based compression strategy VisionZip~\cite{yang2024visionzip}, the pooling-based PLLaVA~\cite{xu2024pllava}, the pre-filling compression method FastV~\cite{chen2024image}, and DyCoke~\cite{tao2024dycoke}, which applies compression in both the temporal and decoding stages. Notably, following the original paper settings, the ratio of dominant to contextual tokens in VisionZip is set to 5.4; the attention computation layer for both FastV and DyCoke is set to layer 3.

\begin{table}[t]
\vspace{-5mm}
    \centering
    \small
    \setlength{\abovecaptionskip}{2pt}
    \definecolor{lightlightgray}{gray}{0.8}
    \tablestyle{5.4pt}{1.0}
    \begin{tabular}{lccccccc}
        \toprule
        Method & Retention ratio &  EgoSchema & MLVU  & VideoMME & VideoMMMU & Avg.(\%)\\
        \midrule
        LLaVA-OV-7B~\cite{li2024llavaone} & 100\% & 58.08 & 62.48 & 57.96 & 40.55 & 100\% \\
        \midrule
        DyCoke~\cite{tao2024dycoke} &50\% & \underline{57.74} & 61.09& 57.35 & 40.22& 98.8\% \\
        PLLaVA~\cite{xu2024pllava}& 50\% & 57.72 & 61.15 & 56.93 & 40.00 & 98.5\% \\
        FastV~\cite{chen2024image}&50\%  & \textbf{58.00}  & \underline{61.27} & \textbf{57.47}&40.44 &99.2\%\\
        VisionZip~\cite{yang2024visionzip}&50\% & 53.57 & 57.03 & 54.19 & 36.89 & 92.0\%\\
        \rowcolor[HTML]{EFEFEF} \textbf{\MyMthd{}}&50\%
         & 57.58 &\textbf{61.32} & \underline{57.37} & \textbf{41.00} & \textbf{99.3\%}\\
        \midrule
        DyCoke~\cite{tao2024dycoke} &35\% & 57.74 & 59.95 & \underline{56.22} & 40.33& \underline{98.0\%}\\
        FastV~\cite{chen2024image}&35\%  &\underline{57.75}  & 59.54 & 56.00 & 39.22& 97.0\%\\
        PLLaVA~\cite{xu2024pllava}& 35\% & 56.07 & 59.42 & 54.26 & 37.44 &94.4\% \\
        VisionZip~\cite{yang2024visionzip}&35\% & 52.00 & 56.29 & 53.70 & 32.78 & 88.3\% \\
        \rowcolor[HTML]{EFEFEF} \textbf{\MyMthd{}}&35\%
       & \textbf{57.94} &\textbf{60.95} & \textbf{57.52} & \textbf{41.33} & \textbf{99.6\%}\\
        \rowcolor[HTML]{EFEFEF} \textbf{\MyMthd{}}&25\%
         & 57.64 &59.81 & \underline{56.44} & \underline{40.33} & 97.9\% \\
        \rowcolor[HTML]{EFEFEF} \textbf{\MyMthd{}}&20\%
         & 57.70 &59.39 & 55.74 & 39.44 & 97.0\%\\
        \midrule
        FastV~\cite{chen2024image}&10\% & 55.87 & 55.81 & 51.63 & 37.11 & 91.5\%\\
        PLLaVA~\cite{xu2024pllava}& 10\% & 53.89 & 54.17 & 50.89 & \underline{38.22} & 90.4\%\\
        VisionZip~\cite{yang2024visionzip}&10\% & 40.78 & 48.42 & 42.56 & 27.56 & 72.3\%\\
        \rowcolor[HTML]{EFEFEF} \textbf{\MyMthd{}}&10\%
         & \textbf{57.52} &\textbf{58.14} & \textbf{55.18} & \textbf{39.11} & \textbf{95.9\%}\\
        \rowcolor[HTML]{EFEFEF}\textbf{\MyMthd{}}&7\%
         & \underline{56.95} &\underline{57.46} & \underline{53.37} & 35.56 & 92.4\%\\
        \rowcolor[HTML]{EFEFEF}\textbf{\MyMthd{}}&5\%
         & 56.61 &56.43 & 53.29 & 36.89 & \underline{92.6}\%\\
        \bottomrule
    \end{tabular}
    \caption{Comparison of state-of-the-art token compression strategies under different token retention ratios on long video understanding benchmarks.}
    \vspace{-20pt}
    \label{tab: long video}
\end{table}

\subsection{Main Results}

\myPara{Results on video question-answering benchmarks.}
\tabref{tab: VQA} presents a comparison of our method with other approaches on video question-answering benchmarks under varying token retention ratios. While all methods achieve performance close to the original model at a 50\% retention ratio, \MyMthd{} shows increasingly advantages as the retention ratio decreases. This highlights the effectiveness of our \MyMthd{} in preserving essential semantic information.

\myPara{Results on long video understanding benchmarks.} Performance on long-video benchmarks are essential for token compression methods. As shown in \tabref{tab: long video}, \MyMthd{} consistently outperforms other methods under the same compression ratios and the advantage of \MyMthd{} becomes increasingly evident under low token budgets. Notably, \MyMthd{} achieves better performance at a 5\% retention ratio than other methods do at 10\%, highlighting its superior efficiency in extreme compression settings.

\myPara{Results on MVBench.}
We also conduct experiments on MVBench~\cite{li2024mvbench}, a comprehensive video understanding benchmark covering 20 tasks organized in the form of multiple-choice questions in \tabref{tab: mvbench}.
\MyMthd{} outperforms all other methods at both 35\% and 10\% retention ratios, demonstrating its consistent superiority in comprehensive scenarios.

\subsection{Ablation Study}
\myPara{Effectiveness of components.} 
In \tabref{component}, we perform ablation experiments on various components of the proposed method. To ensure fairness, we set the retention ratio to 35\% in each experiment. As we can see, when retaining the same number of tokens, the approach using only spatial compression is less effective because redundancy is still present in the tokens, compared to the two-step compression method that eliminates temporal repetition. Additionally, merging all tokens with the selected representative tokens can bring further improvements.

\begin{table}[t]
  \tablestyle{1.8pt}{1.2}
  \setlength{\abovecaptionskip}{2pt}
  \footnotesize
  \scalebox{0.76}{
  \begin{tabular}{lccccccccccccccccccccccc} \toprule
     \textbf{Method} & \textbf{RR} & \textbf{AS} & \textbf{AP} & \textbf{AA} & \textbf{FA} & \textbf{UA} & \textbf{OE} & \textbf{OI} & \textbf{OS} & \textbf{MD} & \textbf{AL} & \textbf{ST} & \textbf{AC} & \textbf{MC} & \textbf{MA} & \textbf{SC} & \textbf{FP} & \textbf{CO} & \textbf{EN} & \textbf{ER} & \textbf{CI} & \textbf{Avg.}\\ 
          \midrule
    LLaVA-OV-7B~\cite{li2024llavaone}  & 100\%
      & 84.0 & 52.0 & 49.0 & 62.0 & 71.0 & 65.0 & 46.5 & 27.0 & 58.5 & 47.5 & 49.0 & 81.0 & 67.5 & 53.0 & 81.0 & 77.5 & 41.5 & 94.5 & 61.5 & 79.5 & 62.43\\ \midrule
    DyCoke~\cite{tao2024dycoke}  & 35\% 
&81.5 & 52.5 & 51.0 & 61.0 & 68.5 & 68.0 & 45.0 & 27.5 & 56.5 & 47.5 & 48.5 & 80.5 & 67.0 & 52.5 & 77.5 & 75.5 & 42.5 & 94.0 & 60.0 & 78.5 & \underline{61.78} \\
    FastV~\cite{chen2024image} & 35\% & 82.0 & 51.0 & 47.5 & 66.0 & 68.0 & 67.5 & 45.0 & 27.0 & 59.0 & 49.0 & 50.0 & 79.0 & 59.0 & 51.5 & 75.5 & 74.5 & 40.5 & 95.0 & 59.5 & 79.0 & 61.28\\
    PLLaVA~\cite{xu2024pllava}& 35\% &81.0 & 45.5 & 50.0 & 59.0 & 64.5 & 63.0 & 43.5 & 27.5 & 57.5 & 45.5 & 40.0 & 77.0 & 64.0 & 53.0 & 77.0 & 69.5 & 39.5 & 94.5 & 60.5 & 77.5 & 59.48\\
    VisionZip~\cite{yang2024visionzip} & 35\% &84.0 & 35.5 & 35.5 & 58.0 & 69.5 & 63.5 & 33.5 & 28.5 & 53.0 & 49.0 & 51.0 & 66.5 & 49.0 & 46.5 & 59.0 & 72.5 & 47.0 & 93.5 & 66.0 & 78.5 & 56.98 \\
    \rowcolor[HTML]{EFEFEF} \textbf{\MyMthd{}} & 35\% & 83.0 & 54.0 & 50.0 & 63.0 & 69.5 & 67.0 & 49.0 & 28.5 & 58.0 & 49.0 & 49.0 & 77.0 & 60.5 & 52.5 & 75.5 & 77.0 & 42.5 & 95.0 & 60.0 & 79.5 & \textbf{61.98}\\
    \midrule
    FastV~\cite{chen2024image} & 10\% & 78.5 & 42.5 & 37.0 & 67.0 & 63.0 & 67.0 & 36.5 & 27.0 & 55.5 & 46.0 & 47.5 & 65.5 & 50.0 & 48.5 & 65.0 & 66.0 & 40.0 & 93.5 & 61.0 & 78.5 & \underline{56.78}\\
    PLLaVA~\cite{xu2024pllava} & 10\% & 77.5 & 37.0 & 42.5 & 54.0 & 63.0 & 60.0 & 37.5 & 28.0 & 51.5 & 46.0 & 35.0 & 70.5 & 56.5 & 48.0 & 73.5 & 64.5 & 39.5 & 93.5 & 60.5 & 78.5 & 55.85 \\
    VisionZip~\cite{yang2024visionzip} & 10\% &
    75.5 & 35.5 & 27.5 & 42.0 & 47.0 & 48.0 & 33.5 & 29.0 & 49.5 & 39.0 & 33.0 & 46.5 & 43.5 & 36.5 & 46.0 & 45.5 & 37.5 & 81.5 & 54.5 & 64.0 & 45.75\\
     \rowcolor[HTML]{EFEFEF} \textbf{\MyMthd{}} & 10\% &   80.5 & 48.0 & 47.5 & 59.0 & 66.5 & 64.5 & 41.0 & 28.0 & 56.5 & 45.0 & 43.0 & 63.0 & 51.0 & 48.0 & 64.5 & 76.0 & 39.5 & 95.5 & 60.5 & 80.0 & \textbf{57.88}\\
     \midrule
    \bottomrule
  \end{tabular}
  }
    \caption{Comparison of state-of-the-art token compression strategies under different token retention ratios on MVBench. `RR' denotes the token retention ratio.
  }\label{tab: mvbench}
  \vspace{-10pt}
\end{table}

\begin{table}[t]
    \setlength{\abovecaptionskip}{-2pt}
    \label{Tab}
    \begin{subtable}{.5\linewidth}
      \centering
      \tablestyle{3pt}{1.0}
      \footnotesize
    \begin{tabular}{ccccccc} \toprule
     Spatial & Temporal & Merge & MVBench & VideoMME \\
    \midrule
    \cmark &        &  &  59.85 & 55.89\\
    \cmark & \cmark &  & 60.55  &55.97\\
    \cmark &        & \cmark &  60.80 &56.40\\
    \cmark & \cmark & \cmark & \textbf{61.98} & \textbf{57.52} \\
    \bottomrule
    \end{tabular}
        \caption{
    Ablation study on different components.
    }
     \label{component}   
    \end{subtable}%
    \begin{subtable}{.49\linewidth}
      \centering
      \tablestyle{5pt}{1.0}
      \footnotesize
    \begin{tabular}{ccccccc} \toprule
    Method & MVBench & VideoMME & ActivityNet \\
    \midrule
    Random & 58.68 & 55.55 & 47.11\\
    Uniform & 61.25 & 56.86 & 47.25\\
    L2Norm & 60.67 & 57.07 & 47.16\\
    \textbf{SCC} & \textbf{61.98} & \textbf{57.52} & \textbf{47.88}\\
    \bottomrule
    \end{tabular}
        \caption{
    Ablation study on token selection methods.
    }
    \label{sample method}
    \end{subtable} 
        \caption{Ablation Studies. `Spatial' and `Temporal' refer to spatial and temporal compression, respectively. `Merge' indicates whether the compressed tokens are merged with all tokens.} 
    \vspace{-20pt}
\end{table}

\myPara{Ablation of token selection methods.} 
To demonstrate the effectiveness of our proposed SCC strategy, we compare it with other methods of selecting representative tokens in \tabref{sample method}. In addition to random sampling and uniform sampling, we also used L2Norm to select tokens with higher information density for comparison. Note that in our experiments, we only replace the method of obtaining representative tokens, while the `Merge'
operation in \tabref{component} is retained for a fair comparison. It is evident that our method based on SCC performs better than other sampling methods.

\begin{figure}[tp!]
      \setlength{\abovecaptionskip}{2pt}
     \centering
         \begin{subfigure}[b]{0.32\textwidth}
         \centering
         \includegraphics[width=\textwidth]{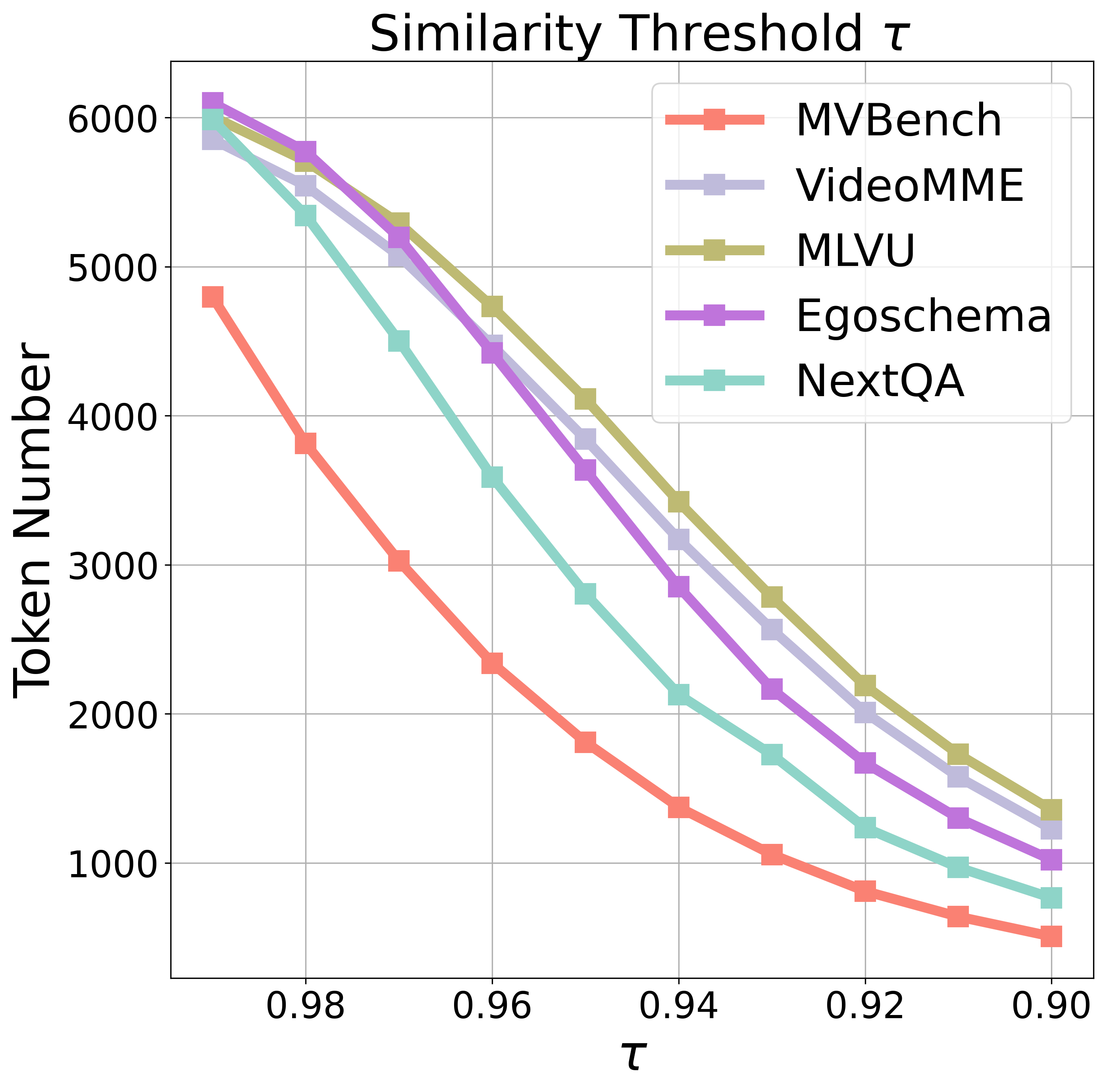 }
         \caption{Impact of $\tau \in [0.9, 0.99]$  }
         \label{fig: tau 1}
     \end{subfigure}
     \hfill
    \begin{subfigure}[b]{0.32\textwidth}
         \centering
         \includegraphics[width=\textwidth]{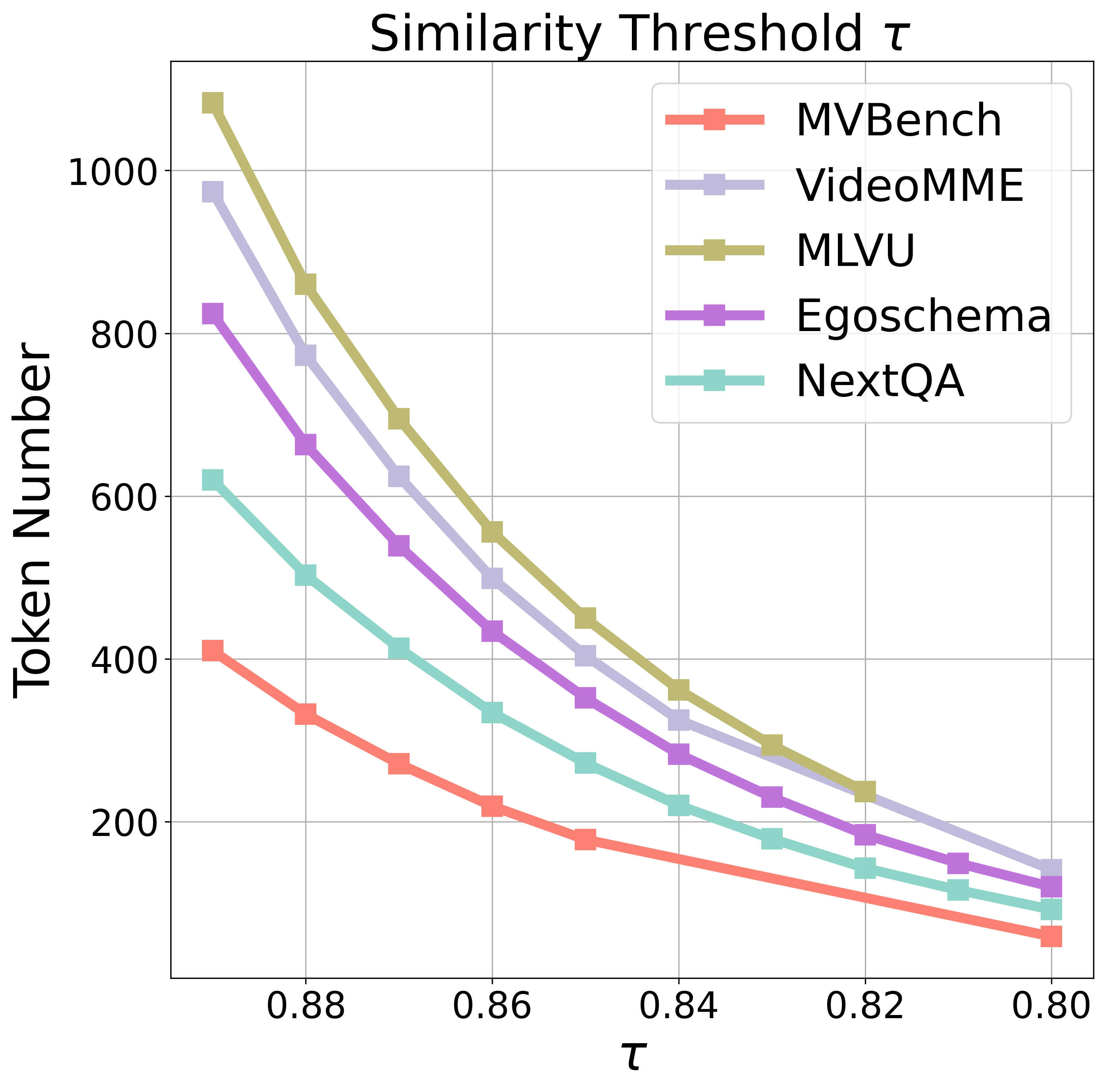}
         \caption{Impact of $\tau \in [0.8, 0.89]$ }
         \label{fig: tau 2}
     \end{subfigure}
          \begin{subfigure}[b]{0.32\textwidth}
         \centering
         \includegraphics[width=\textwidth]{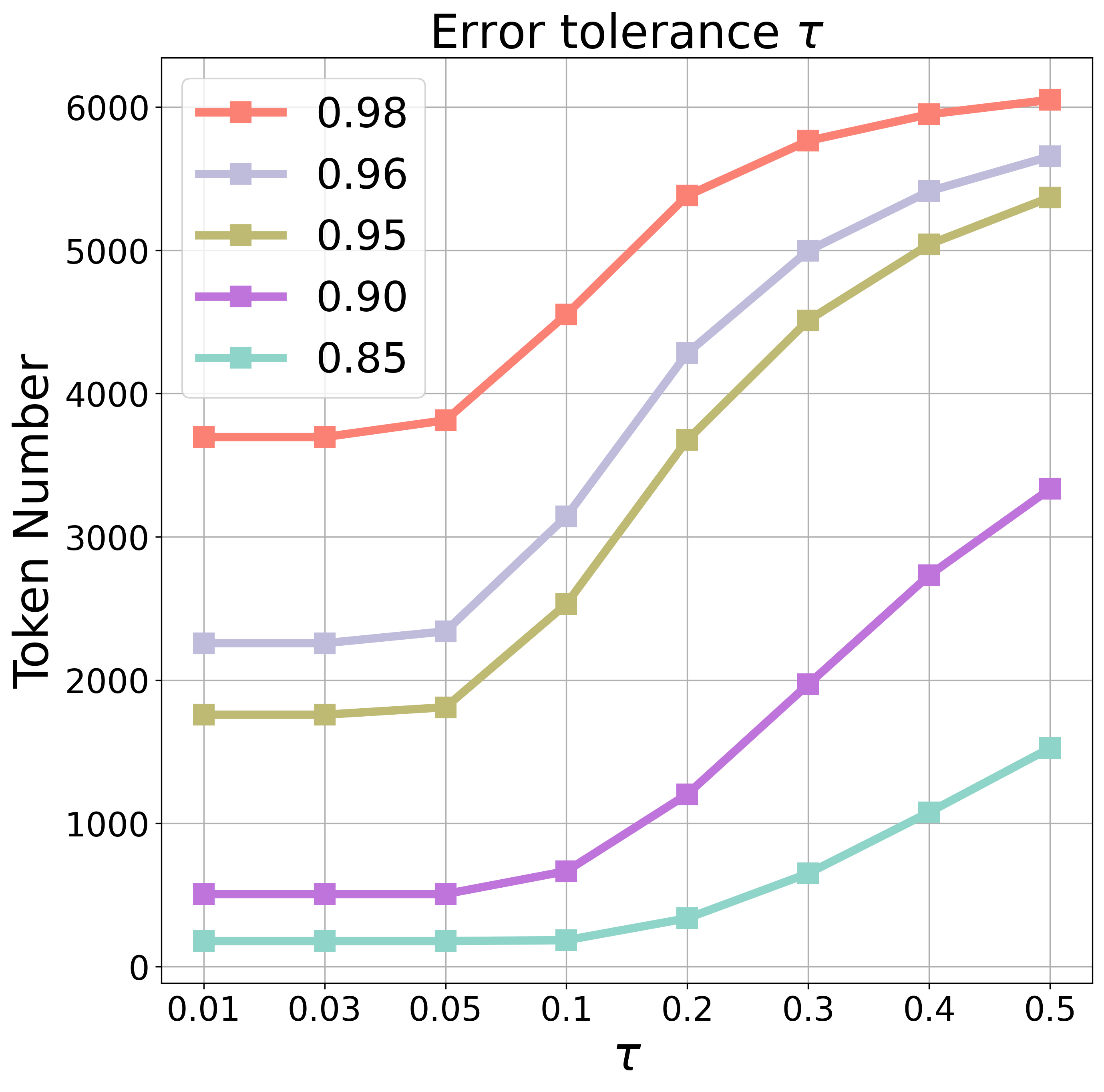}
         \caption{Impact of $\epsilon$}
         \label{fig: epsilon}
     \end{subfigure}
     \hfill

        \caption{Token number statistics of similarity threshold $\tau$ and error tolerance $\epsilon$.}
        \label{fig:statistics}
\end{figure}

\myPara{Impact of similarity threshold $\tau$ and error tolerance $\epsilon$.}
The similarity threshold $\tau$ and error tolerance $\epsilon$ are the key parameters in controlling the accuracy of the connectivity graph, as well as the core variables for managing the token number after compression. In \figref{fig: tau 1} and \figref{fig: tau 2}, we discuss how the similarity threshold $\tau$ affects the number of tokens after compression across different benchmarks. \figref{fig: tau 1} corresponds to $\tau$ values ranging from 0.99 to 0.9, while Figure \figref{fig: tau 2} covers $\tau$ values from 0.89 to 0.8. It is evident that as the threshold for inclusion in connected components becomes more lenient, more tokens are compressed. Additionally, due to the varying distributions of videos across different benchmarks, the number of tokens under the same threshold $\tau$ varies.

The error tolerance $\epsilon$  in \eqnref{eqn: sample} serves as a trade-off between the accuracy of solving connected components and the computational efficiency. Theoretically, the lower the $\epsilon$, the more precise the connected components obtained. In Figure \ref{fig: epsilon}, we illustrate how token numbers vary with changes in $\epsilon$ under different $\tau$ values on MVBench~\cite{li2024mvbench}. Since we treat uncovered tokens as separate connected components, as $\epsilon$ increases, the token number also tends to increase. 
It can be observed that when $\epsilon$ is smaller than 0.05, the connected components almost no longer change. This indicates that nearly all distinct semantic regions have been identified and merged. Therefore, we set  $\epsilon = 0.05$.

\begin{table}[t]
    \begin{subtable}{.5\linewidth}
      \centering
      \tablestyle{3.2pt}{1.1}
      \footnotesize
    \begin{tabular}{lcccccc} \toprule
    Method & RR & FLOPs & FR & Avg. \\
    \midrule
    LLaVA-OV-7B~\cite{li2024llavaone} & 100\% & 41.4 & 100\% & 100\%\\
    Dycoke~\cite{tao2024dycoke} &50\%& 18.7 & 45.2\% & 99.2\%\\
    FastV~\cite{chen2024image} & 50\% & 21.4 & 51.7\% & 99.2\%\\
    \rowcolor[HTML]{EFEFEF} \MyMthd{} & 50\% & 18.6 & 44.9\% & 99.5\%\\
    Dycoke~\cite{tao2024dycoke} &35\%& 13.1 & 31.6\% & 98.4\%\\
    FastV~\cite{chen2024image} & 35\% & 16.1 & 38.9\% & 98.3\%\\
    \rowcolor[HTML]{EFEFEF} \MyMthd{} & 35\% & 13.4 & 32.4\% & 99.3\%\\
    \bottomrule
    \end{tabular}
        
    \end{subtable}%
    \begin{subtable}{.5\linewidth}
      \centering
      \tablestyle{4.8pt}{1.1}
      \footnotesize
    \begin{tabular}{lcccccc} \toprule
    Method & RR & FLOPs & FR & Avg. \\
    \midrule
    \rowcolor[HTML]{EFEFEF} \MyMthd{} & 20\% & 7.6 & 18.3\% & 98.3\%\\
    VisionZip~\cite{yang2024visionzip} &10\% &3.9 & 9.42\% & 78.8\%\\
     PLLaVA~\cite{xu2024pllava}&10\%& 3.9 & 9.42\% & 92.7\% \\
    FastV~\cite{chen2024image} & 10\% & 8.1 & 19.6\% & 93.1\% \\
    \rowcolor[HTML]{EFEFEF} \MyMthd{} & 10\% & 4.0 &9.66\% & 96.6\%\\
    \rowcolor[HTML]{EFEFEF} \MyMthd{} & 7\% & 2.9 & 6.28\% & 94.7\%\\
    \rowcolor[HTML]{EFEFEF} \MyMthd{} & 5\% & 2.3 & 5.56\%& 94.5\%\\
    \bottomrule
    \end{tabular}
    \end{subtable} 
    \caption{Comparison of FLOPs and Performance across Methods.
`RR' refers to the Retention Ratio, `FR' refers to the FLOPs Ratio, and `Avg.' denotes the average performance ratio.} 
            \label{flops}
    \vspace{-5pt}
\end{table}

\subsection{ Efficiency Analysis}
Following Dycoke~\cite{tao2024dycoke}, we divide the FLOPs generated during the LLM stage into two parts: the prefilling stage and the decoding stage. 
In the prefilling stage, for each transformer layer, the computational cost of the multi-head attention and the FFN can be expressed as $4kd^2 + 2k^2d + 2kdc $, where $k, d$, and $c$ denote the number of tokens, the hidden state size, and the intermediate size of the FFN, respectively. In the decoding stage, thanks to the KV cache, the computational cost for decoding each token is significantly reduced and can be represented as $(4d^2 + 2dc) + 2(dk + \frac{d(R+1)}{2})$. 
Therefore, the total FLOPs during the LLM stage can be expressed as:  
\begin{equation}
    \text{FLOPs} = T*(4kd^2+2k^2d+2kdc) + TR((4d^2+2dc)+2(dk+\frac{d(R+1)}{2})),
\end{equation}
where $T$, $R$ represents the number of transformer layers and predicted token length. We set $R=100$ in all calculations. Additionally, we further calculate the FLOPs introduced by our \MyMthd{} during token compression. Since the main computational cost stems from similarity computation~\cite{sun2023corrmatch,yang2025topv,shen2024longvu} between tokens, the FLOPs generated by \MyMthd{} can be expressed as:
\begin{equation}
    \text{FLOPs}' = n* 2m^2d + 2k_1^2d + 2 nmk_2d,
\end{equation}
where $n, m $, and $d$ represent the number of frames, the number of tokens per frame, and the hidden state size, respectively. 
$k_1$ and $k_2$ represent the number of tokens after the first step of spatial compression and the second step of temporal compression, respectively.

In \tabref{flops}, we compare the FLOPs and average performance across all benchmarks of our method with other methods under different retention ratios. Unlike FastV~\cite{chen2024image} that performs compression at the LLM stage, our \MyMthd{} compresses tokens before LLMs, significantly reducing FLOPs. Furthermore, it can be observed that our method consistently achieves the highest average performance at the same retention ratio, especially when the retention ratio is relatively low. Our method demonstrates an advantage over methods like PLLaVA~\cite{xu2024pllava} and VisionZip~\cite{yang2024visionzip}.

\section{Reducing Law in Token Compression}
\label{sec: reducing law}
Observing \tabref{tab: VQA}, \tabref{tab: long video}, and \tabref{tab: mvbench}, there is a common phenomenon: When the retention ratio is relatively high, existing methods, including \MyMthd{}, can maintain model performance quite well. However, as the retention ratio drops, the model performance begins to decline rapidly. To further analyze the impact of the retained token numbers on performance, here we attempt to analyze the reducing law in video token compression.

\subsection{Token Redundancy in Video MLLMs }
\begin{figure}
      \setlength{\abovecaptionskip}{2pt}
     \centering
         \begin{subfigure}[b]{0.245\textwidth}
         \centering
         \includegraphics[width=\textwidth]{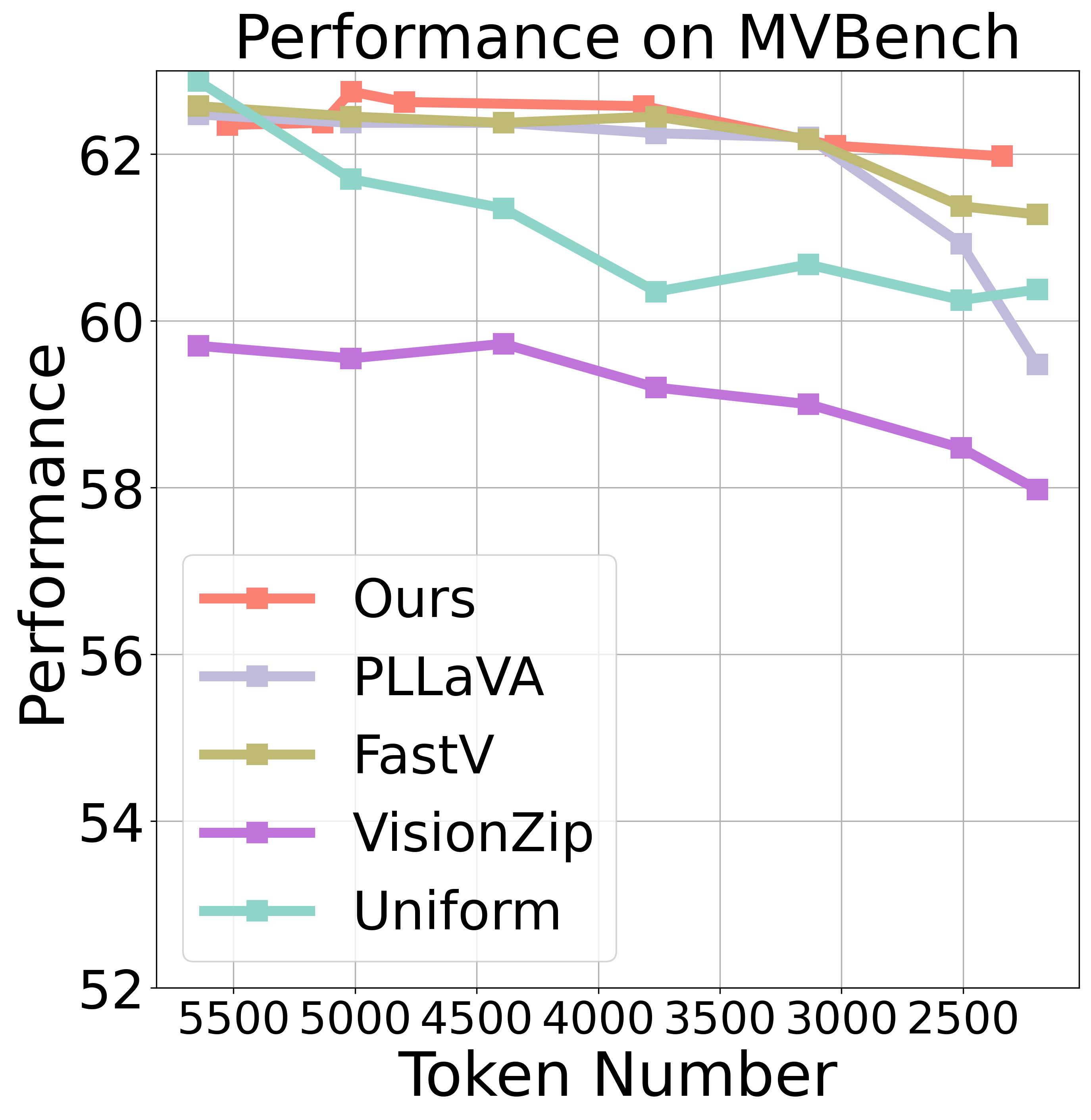}
         \caption{High RR on MVBench}
         \label{fig: mvbench high}
     \end{subfigure}
     \hfill
    \begin{subfigure}[b]{0.245\textwidth}
         \centering
         \includegraphics[width=\textwidth]{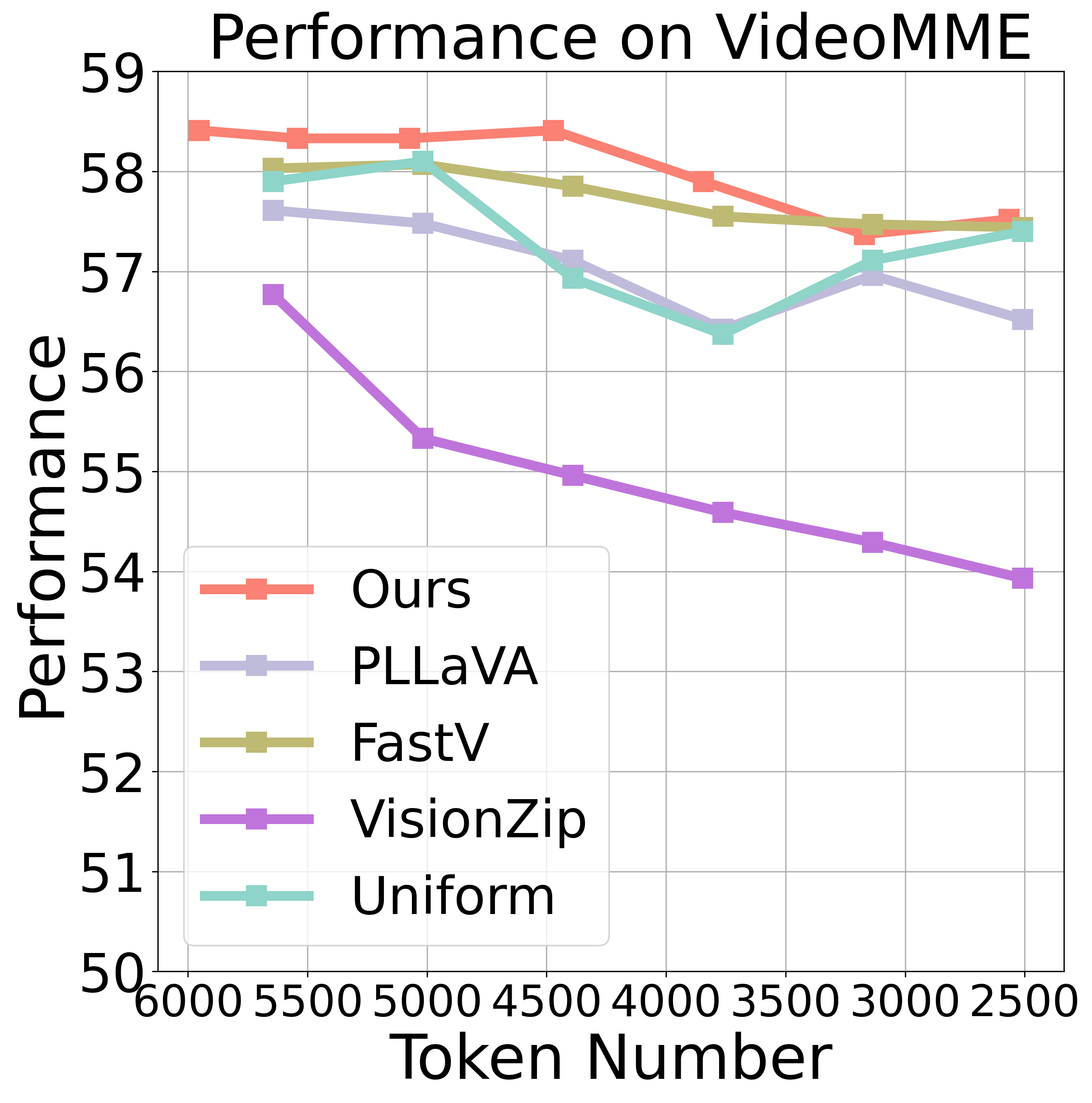}
         \caption{High RR on VideoMME }
         \label{fig: videomme high}
     \end{subfigure}
          \begin{subfigure}[b]{0.245\textwidth}
         \centering
         \includegraphics[width=\textwidth]{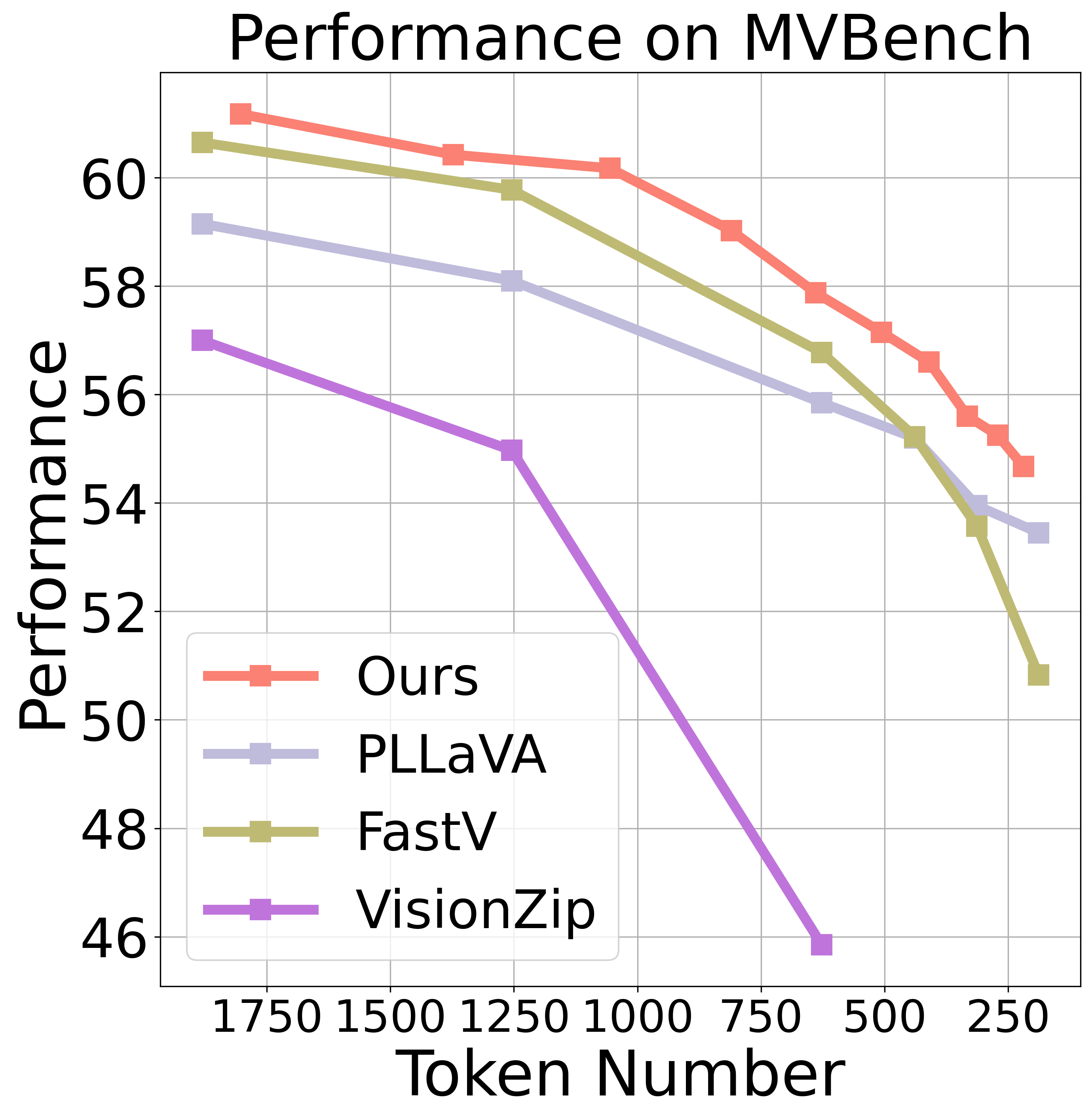}
         \caption{Low RR on MVBench}
         \label{fig: mvbench low}
     \end{subfigure}
     \hfill
    \begin{subfigure}[b]{0.245\textwidth}
         \centering
         \includegraphics[width=\textwidth]{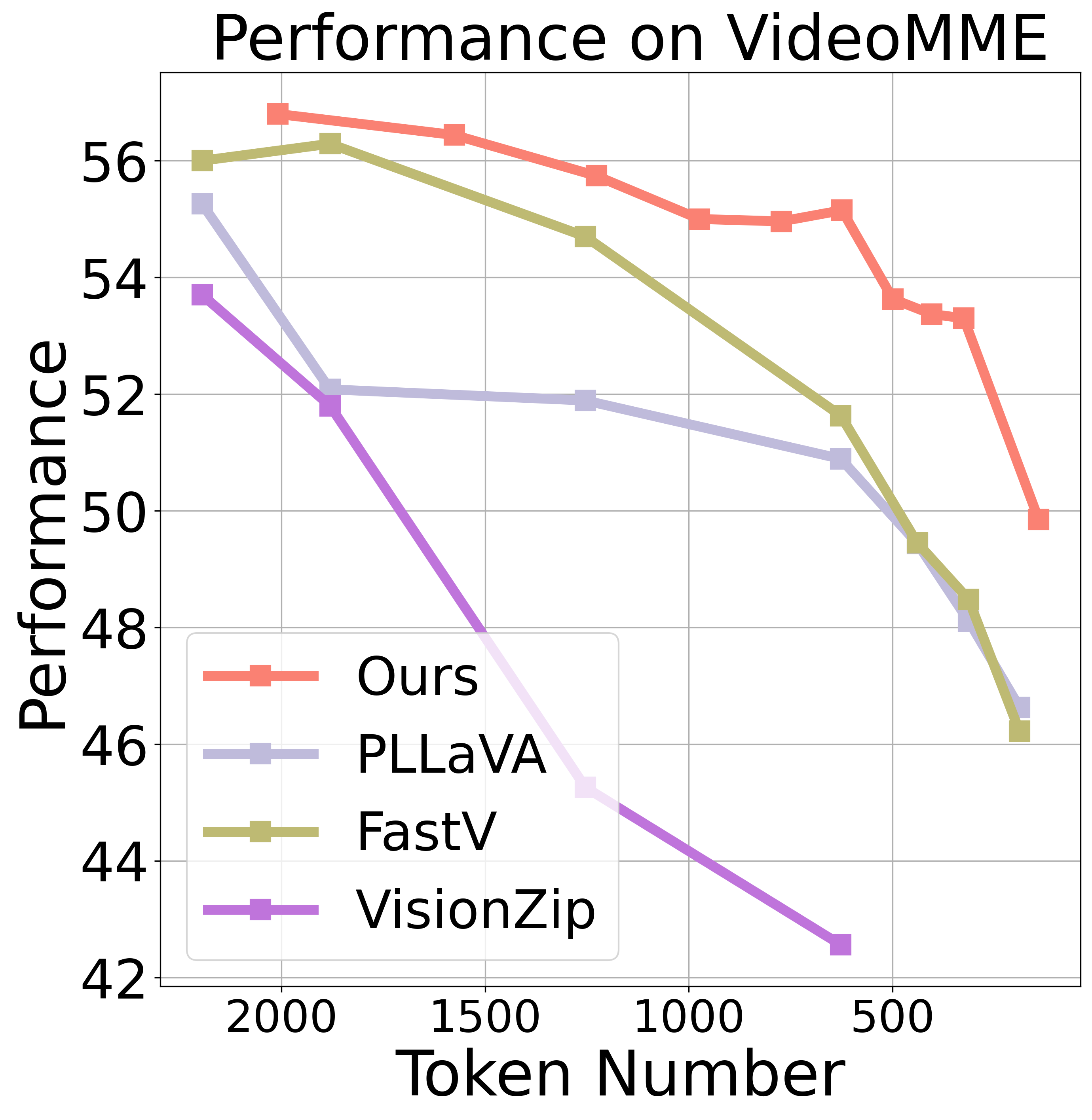}
         \caption{Low RR on VideoMME}
         \label{fig: videomme low}
     \end{subfigure}
     \hfill
         \vspace{-10pt}
        \caption{Performance degradation of methods on different benchmarks as the retained token number decreases. `RR' denotes the token retention ratio.}
        \label{fig: reducing law low}
         \vspace{-5pt}
\end{figure}

In \figref{fig: mvbench high} and \figref{fig: videomme high}, we present a comparative analysis of different models (PLLaVA~\cite{xu2024pllava}, FastV~\cite{chen2024image}, VisionZip~\cite{yang2024visionzip}, and \MyMthd{}) performance on two benchmarks: MVBench~\cite{li2024mvbench} and VideoMME~\cite{fu2024videomme}, in a range of token retention ratios from 90\% to 35\%  (with token number from 5644 to 2195). The results demonstrate that most token reduction methods achieve performance that remains comparable to the original model. 
Notably, even the naive uniform sampling strategy introduces only minimal degradation in performance, which suggests a high degree of redundancy among visual tokens in VLLMs. 

This observation implies that many tokens contribute little to the final prediction and can be safely discarded or aggregated without significantly affecting model output. 
Redundancy is common in VLLMs, as similar content within and across frames often leads to duplicated visual tokens.
These findings provide strong empirical support for developing more efficient token selection and compression strategies without sacrificing too much accuracy.


\subsection{Semantic Loss under Aggressive Token Reduction}
\label{sec: low ratio}
As shown in \figref{fig: mvbench low} and \figref{fig: videomme low}, we evaluate the performance degradation of different token selection methods on the MVBench~\cite{li2024mvbench} and VideoMME~\cite{fu2024videomme} benchmarks, under aggressive token retention ratios (35\% to 3\%, with token number range from 2195 to 181). In contrast to the results in \figref{fig: mvbench high} and \figref{fig: videomme high}, where most methods maintain performance levels comparable to the original model, we observe a sharp performance decline for all methods as the retention ratio drops below 35\%. This indicates that the indiscriminate reduction or aggregation of tokens at very low budgets results in the loss of semantically critical visual information, which significantly hampers model understanding.

It is evident that when retaining a smaller number of tokens, our \MyMthd{} demonstrates consistent superiority over other methods.
Specifically, at a 3\% retention ratio, the average scores across the two benchmarks of FastV~\cite{chen2024image} drop to 80.7\% (81.4\% on MVBench and 79.7\% on VideoMME), whereas \MyMthd{} retains a substantially higher score of 86.8\% (+6.1\%). The performance gaps highlight the advantage of \MyMthd{}, which prioritizes semantically representative and diverse tokens while minimizing redundancy. As a result, our method is better suited to tight token budgets, preserving the essential visual semantics needed for accurate video-language understanding.

\section{Conclusions}
In this paper, we propose \MyMthd{}, a training-free token compression paradigm for Video Large Language Models. We first introduce the SCC strategy, which achieves token compression by identifying semantic connected components within the token set. By applying the SCC strategy in both spatial and temporal domains, \MyMthd{} uses a set of non-overlapping tokens to represent the entire video, effectively eliminating spatio-temporal redundancy in video tokens. Experimental results and an analysis of the reducing law in video compression demonstrate that \MyMthd{} effectively preserves key semantic information during video token compression, especially when the number of retained tokens is low.


{
    \small
    \bibliographystyle{ieeenat_fullname}
    \bibliography{main}
}

\newpage
\appendix

\section*{Appendix}

In order to provide a more complete illustration of the capabilities of \MyMthd{}, a comprehensive appendix has been developed. This appendix includes detailed descriptions of the evaluation benchmarks, additional experiments results,  algorithms analysis, and discussion among limitations and broader impacts.
\hypersetup{linkbordercolor=Black,linkcolor=Black}
\etoctoccontentsline{part}{Appendix}
\localtableofcontents
\hypersetup{linkbordercolor=BrickRed,linkcolor=BrickRed}
\section{Benchmarks}
\label{sec: benchmarks}
\subsection{Video Question-Answering Benchmarks}
Video question-answering is a fundamental capability of Video Large Language Models (VLLMs). In this study, we evaluate \MyMthd{}’s performance on this task using widely adopted benchmarks, including ActivityNet-QA~\cite{caba2015activitynet}, VideoChatGPT~\cite{Maaz2023VideoChatGPT}, and Next-QA~\cite{xiao2021next}.

ActivityNet-QA~\cite{caba2015activitynet} consists of human-annotated, action-related question-answer pairs derived from the ActivityNet dataset, with an average video duration of approximately 2 minutes. The VideoChatGPT~\cite{Maaz2023VideoChatGPT} benchmark evaluates five critical aspects of video understanding: accuracy of information, detail orientation, contextual comprehension, temporal reasoning, and consistency. Next-QA~\cite{xiao2021next} focuses on reasoning about causal and temporal actions, as well as understanding rich object interactions in daily activities.

\subsection{Long Video Understanding Benchmarks}
Handling long videos is crucial for evaluating token compression strategies because longer sequences pose greater challenges in balancing token efficiency and preserving essential information. To demonstrate that \MyMthd{} can effectively manage diverse video scenarios, we evaluate it on several long video understanding benchmarks, including EgoSchema~\cite{mangalam2024egoschema}, MLVU~\cite{zhou2024mlvu}, VideoMME~\cite{fu2024videomme}, and VideoMMMU~\cite{hu2025videommmu}.

Among these benchmarks, EgoSchema\cite{mangalam2024egoschema} comprises egocentric videos captured from a first-person perspective, with an average duration of approximately 180 seconds. MLVU\cite{zhou2024mlvu} emphasizes understanding extended videos with lengths varying widely from 3 minutes up to 2 hours, pushing models to maintain temporal coherence and effectively summarize long-term dependencies. VideoMME\cite{fu2024videomme} includes videos from diverse domains with durations ranging from minutes to hours, making it one of the most comprehensive and challenging benchmarks for holistic video understanding. Finally, Video-MMMU\cite{hu2025videommmu} serves as a multi-modal and multi-disciplinary benchmark that evaluates large multimodal models on their ability to not only comprehend but also integrate and apply knowledge from videos across various fields, thereby testing cross-domain reasoning and knowledge transfer capabilities.
\subsection{MVBench}
In addition to the previously discussed video question-answering and long video understanding benchmarks, we also evaluate our method on MVBench~\cite{li2024mvbench}, a comprehensive and challenging mluti-choice video understanding benchmark. MVBench is designed to assess the temporal comprehension capabilities by presenting 20 distinct tasks in the form of multiple-choice questions. These tasks cover diverse scenarios that require sophisticated temporal reasoning and understanding of dynamic content, which cannot be achieved through single-frame analysis alone.
\section{Supplementary Experimental Analysis}
\subsection{More comparison of existing methods}
We further extend our evaluation by comparing \MyMthd{} with a wider range of token compression strategies, as shown in \tabref{tab: benchmark supp}. Notably, these approaches are implemented upon different baseline models, such as LLaVA-OneVision~\cite{li2024llavaone} and Video-LLaVA~\cite{lin2023video}, which introduces inherent discrepancies in absolute performance. To ensure a fair and consistent comparison of compression efficiency, we report the relative performance with respect to each method’s own baseline across various token retention ratios. The experimental results demonstrate that \MyMthd{} consistently outperforms other approaches at both moderate (35\%) and aggressive (10\%) token retention ratios, underscoring its robustness and superior capability in preserving critical semantic information under constrained token budgets.
\begin{table}[t]
    \centering
    \small
    \setlength{\abovecaptionskip}{2pt}
    \definecolor{lightlightgray}{gray}{0.8}
    \tablestyle{4pt}{1.0}
    \begin{tabular}{lccccccc}
        \toprule
        Method & Retention ratio &  ActivityNet & MVBench & MLVU  & VideoMME & NextQA & Avg.(\%)\\
        \midrule
        PruMerge~\cite{shang2024LLaVAPruMerge} & 55\% &97.2\% & 90.7\% & - & 90.3\% & 95.6\% & 93.5\%\\
        DyCoke~\cite{tao2024dycoke} &35\% & 99.4\% & 99.0\%& 96.0\% & 97.0\%& 99.4\% & 98.2\%\\
        PLLaVA~\cite{xu2024pllava}& 35\% & 98.2\% & 95.3\% & 95.1\% & 93.6\% & 97.9\% & 96.0\% \\
        FastV~\cite{chen2024image}&35\%  &99.5\% & 98.2\%  & 95.3\% & 96.6\%&99.0\% & 97.7\%\\
        VisionZip~\cite{yang2024visionzip}&35\% & 94.4\% & 91.3\% & 90.1\% & 92.7\% & 95.6\%& 92.8\%\\
        PACT~\cite{dhouib2025pactpruningclusteringbasedtoken} & 35\% & 98.9\% & - & \textbf{99.2\%} & 98.4\% & - & 98.3\% \\
        FrameFusion~\cite{fu2024framefusion} & 30\% & - & - & - & 96.7\% & 98.3\% & 97.5\% \\
        \rowcolor[HTML]{EFEFEF} \textbf{\MyMthd{}}&35\%
         & \textbf{99.6\%} &\textbf{99.3\%} & \underline{97.6\%} & \textbf{99.2\%} & \textbf{99.4\%} & \textbf{99.0\%}\\
         \midrule

        DiVPrune~\cite{divprune} & 15\% & 95.4\% & - & - & - & 95.1\% & 95.3\% \\
        VisionZip~\cite{yang2024visionzip}&10\% & 80.2\% & 73.3\% &77.5\% &73.4\% &  80.0\% & 76.9\%\\
        PLLaVA~\cite{tao2024dycoke} & 10\% & 94.1\% & 89.5\% & 86.7\% &87.8\%\ & 95.7\% & 90.8\%\\
        FastV~\cite{chen2024image}&10\% & 93.5\% & 90.9\% & 89.3\% & 89.1\% & 96.8\% & 91.9\%\\
        PruneVID~\cite{huang2024prunevid} & 10\% & - & 89.9\% & 87.9\% & 88.0\% & - &  88.6\%\\
        \rowcolor[HTML]{EFEFEF} \textbf{\MyMthd{}}&10\%
         & \textbf{99.3\%} &\textbf{92.7\%} & \textbf{93.1\%} & \textbf{95.2\%} & \textbf{98.4\%} & \textbf{95.7\%}\\

        \bottomrule
    \end{tabular}
    \caption{Comparison of state-of-the-art token compression strategies under different token retention ratios on various video understanding benchmarks.}
    \label{tab: benchmark supp}
\end{table}

\subsection{Results with LLaVA-OneVision 0.5B model}
Given that token compression strategies are often used in resource-constrained scenarios, it is particularly valuable to investigate their effectiveness when applied to smaller base models. Unlike large-scale models with abundant parameters and computational capacity to compensate for potential information loss during compression, lightweight models are inherently limited in their ability to recover or reason over missing information. Therefore, preserving more essential semantic content becomes even more critical.

To evaluate this, we deploy \MyMthd{} in LLaVA-OneVision-0.5B base model (enhanced with SIGLIP~\cite{zhai2023sigmoid} as encoder and Qwen-2.5-0.5B~\cite{qwen2.5} as LLM), which contains significantly fewer parameters. As shown in \tabref{tab: 0.5b supp}, \MyMthd{} demonstrates performance patterns consistent with those observed in the 7B model setting. Despite the significantly reduced model capacity, our method effectively preserves key semantic representations and mitigates the performance drop commonly associated with token compression. In comparison with FastV, \MyMthd{} consistently outperforms it under both 35\% and 10\% token retention ratios.
These findings underscore the robustness and practical utility of our method in real-world applications, particularly for deployment on edge devices or mobile platforms where both model size and computational budgets are limited.

\begin{table}[t]
    \centering
    \small
    \setlength{\abovecaptionskip}{2pt}
    \definecolor{lightlightgray}{gray}{0.8}
    \tablestyle{4pt}{1.0}
    \begin{tabular}{lccccccc}
        \toprule
        Method & RR &  VideoChatGPT & MVBench & MLVU  & VideoMME & Egoschema & Avg.(\%)\\
        \midrule
        LLaVA-OV-0.5B~\cite{li2024llavaone} & 100\% & 2.89 & 51.23 & 46.78 & 40.15 &40.71 & 100\% \\
        \midrule
        \MyMthd{} &50\% & 2.84 & 51.28& 46.99 & 39.93& 41.11 & 99.9\% \\
        \midrule
        FastV~\cite{chen2024image}&35\%  &2.78  & 50.80 & 46.45&39.25 &40.74 & 98.5\%\\
        \MyMthd{} & 35\% &\textbf{2.79} & \textbf{51.03} & \textbf{47.39} & \textbf{39.37} & \textbf{41.13} & \textbf{99.3\%}\\
        \midrule
        FastV~\cite{chen2024image}&10\%  &2.70  & 50.10 & -&36.14 &38.38 & 93.9\%\\
        \MyMthd{} & 10\% & \textbf{2.72 }& \textbf{50.38} & \textbf{46.62} & \textbf{38.48} & \textbf{40.79} & \textbf{97.6\%}\\
        \MyMthd{} & 7\% & 2.69& 49.95 & 45.99 & 38.22 &   40.17 & 96.6\%\\
        \MyMthd{}& 5\% & 2.67 & 49.60 & 45.91 & 37.89 &  39.57 & 95.8\% \\
        \bottomrule
    \end{tabular}
    \caption{Evaluation of \MyMthd{} and FastV~\cite{chen2024image} on various video understanding benchmarks with the LLaVA-OneVision-0.5B base model under varying token retention ratios.}
    \vspace{-10pt}
    \label{tab: 0.5b supp}
\end{table}

\begin{algorithm}[t]
    \caption{Union-Find Data Structure with Path Compression and Union by Rank}
    \label{alg:unionfind}
    \begin{algorithmic}[1]
        
        \State Initialize $\texttt{parent}[i] = i$ for all $i$
        \State Initialize $\texttt{rank}[i] = 0$ for all $i$

        \Function{find}{$x$}
            \While{$\texttt{parent}[x] \neq x$}
                \State $\texttt{parent}[x] \gets \texttt{parent}[\texttt{parent}[x]]$ \Comment{Path compression}
                \State $x \gets \texttt{parent}[x]$
            \EndWhile
            
            \Return $x$
        \EndFunction

        \Function{batch\_union}{$\texttt{x\_arr}, \texttt{y\_arr}$}
            \For{$(x, y) \in \texttt{zip}(\texttt{x\_arr}, \texttt{y\_arr})$}
                \State $x_{\text{root}} \gets \textbf{find}(x)$
                \State $y_{\text{root}} \gets \textbf{find}(y)$
                \If{$x_{\text{root}} = y_{\text{root}}$}
                    \State \textbf{continue}
                \EndIf
                \If{$\texttt{rank}[x_{\text{root}}] < \texttt{rank}[y_{\text{root}}]$}
                    \State $\texttt{parent}[x_{\text{root}}] \gets y_{\text{root}}$
                \Else
                    \State $\texttt{parent}[y_{\text{root}}] \gets x_{\text{root}}$
                    \If{$\texttt{rank}[x_{\text{root}}] = \texttt{rank}[y_{\text{root}}]$}
                        \State $\texttt{rank}[x_{\text{root}}] \gets \texttt{rank}[x_{\text{root}}] + 1$
                    \EndIf
                \EndIf
            \EndFor
        \EndFunction
    \end{algorithmic}
\end{algorithm}

\section{Algorithms Analysis}
\label{sec: algorithms}
\subsection{Union-Find Data Structure}
The Union-Find is an efficient data structure designed to manage and merge disjoint sets. It supports two fundamental operations: the `Find' operation to determine the root of the set for a particular element and the `Union' operation to merge two disjoint sets into one.

In practice, the Union-Find data structure is typically optimized using path compression and union-by-rank. Specifically, path compression flattens the tree structure during the `Find' operation by directly linking all nodes along the query path to the root, thereby reducing the time complexity of subsequent queries. Union-by-rank, on the other hand, attaches the smaller tree to the root of the larger tree during the `Union' operation, preventing excessive growth in tree height. We provide the detailed the Union-Find structure in \agref{alg:unionfind}.
\label{sec: unionfind}

\subsection{Approximate Connected Components Algorithm}
\label{sec: approximate components}
Equipped with the Union-Find data structure, we further leverage it to compute the connected components of a graph, as detailed in \agref{alg:approximate_components}. Specifically, given a graph with $\textbf{N}$ nodes, our approximate connected components algorithm begins by initializing the Union-Find structure with each node as an individual set. We then sample $\textbf{N}'$ nodes from the graph to construct an adjacency list and mark each corresponding edge. For each edge in this sampled graph, a Union operation is performed to iteratively merge connected components. After all edges have been processed, we traverse all nodes and invoke the Find operation to determine the root set to which each node belongs, thereby identifying the final connected components. It is important to note that nodes not covered by the sampled connections are treated as single connected component. Finally, considering the relative positional relationships between tokens represented by the nodes, we sort the connected components based on the node with the highest degree within each cluster.
\begin{algorithm}[t]
    \caption{Approximate Connected Components via Union-Find}
    \label{alg:approximate_components}
    \begin{algorithmic}[1]
        \Require Adjacency matrix $\texttt{adj\_matrix}$, error tolerance $\epsilon$
        \Ensure List of connected components
        
        \State $n \gets \texttt{adj\_matrix.shape[0]}$
        \State Initialize $\texttt{nodes\_flag}[i] = 1$ for all $i$
        \State $\texttt{uf} \gets UnionFind(n)$  \Comment{Create an Union-Find set with n nodes}
        
        \State $\texttt{sample\_size} \gets \min(n, \lceil \log(n) / \epsilon^2 \rceil)$
        \State $\texttt{sampled\_nodes} \gets \texttt{random.sample}(\{1, 2, ..., n\}, \texttt{sample\_size})$
        \State Set $\texttt{nodes\_flag}[i] = 0$ for all $i \in \texttt{sampled\_nodes}$

        \State $\texttt{neighbor\_dict} \gets [] $ \Comment{ Initialize adjacency list for $\texttt{sampled\_nodes}$}
        \For{$i \in \texttt{sampled\_nodes}$}         
            \State $\texttt{neighbors} \gets \{j \mid \texttt{adj\_matrix}[i][j] \neq 0\}$ 
            \State $\texttt{neighbor\_dict}[i] \gets \texttt{neighbors}$
            \State Set $\texttt{nodes\_flag}[j] = 0$ for all $j \in \texttt{neighbors}$
        \EndFor

        \State $\texttt{all\_x}, \texttt{all\_y} \gets [], []$
        \For{$i \in \texttt{sampled\_nodes}$}
            \For{$j \in \texttt{neighbor\_dict}[i]$}
                \State $\texttt{all\_x.append}(i)$
                \State $\texttt{all\_y.append}(j)$
            \EndFor
        \EndFor
        \State $\texttt{uf.batch\_union}(\texttt{all\_x}, \texttt{all\_y})$ \Comment{Merge subgraphs of all \texttt{sampled\_nodes} }

        \State $\texttt{sampled\_roots} \gets [\texttt{uf.find}(i) \text{ for } i \in \texttt{sampled\_nodes}]$ \Comment{Find root for \texttt{sampled\_nodes}}
        \State $\texttt{unique\_roots} \gets \texttt{Unique}(\texttt{sampled\_roots})$ \Comment{Find every unique connected component root}

        \State $\texttt{components} \gets []$
        \For{$\texttt{root} \in \texttt{unique\_roots}$}
            \State $\texttt{cluster} \gets \texttt{np.where}(\texttt{uf.parent} == \texttt{root})[0].\texttt{tolist}()$
            \State $\texttt{components.append}(\texttt{cluster})$

        \EndFor
        \State $\texttt{remain\_nodes} \gets \{i \mid \texttt{nodes\_flag}[i] = 1\}$ \Comment{In case there are any unconsidered nodes.}
        \State $\texttt{components.extend}(\texttt{remain\_nodes})$ \Comment{Treat unconsidered nodes as independent components}

        \State $\texttt{components.sort()}$ \Comment{Sort by the node ID with the highest degree in each cluster. }

    \end{algorithmic}
\end{algorithm}


\subsection{Complexity Analysis}
We first analyze the computational complexity of finding connected components using the Union-Find algorithm without approximation, i.e., when all nodes are considered.  For the Union-Find structure optimized with path compression and union by rank, the `Find' and `Union' operations have an amortized time complexity of $O(\alpha(n))$, where $\alpha(\cdot)$ denotes the inverse Ackermann function, which is generally regarded as a constant less than 5 in practice~\cite{1985unionfind,2005unionfind}. The initialization of the Union-Find data structure takes $O(n)$ time, and performing Union on each of the $m$ edges results in an overall complexity of $O(m\alpha(n))$. Therefore, for a graph with $n$ nodes and $m$ edges, the total complexity of identifying connected components using Union-Find is $O(n + m\alpha(n))$.

In our case, since the graph is represented using an adjacency matrix to encode pairwise similarity among tokens, the number of edges in the worst case is $\mathbf{N}^2$, where $\mathbf{N}$ is the total number of tokens. Thus, without applying any approximation or sparsification, the overall worst-case complexity for computing connected components becomes $O(\mathbf{N}^2\alpha(\mathbf{N}))$.

To further improve computational efficiency, we propose an approximation strategy, in which $\mathbf{N}' = \min(\mathbf{N}, \lceil \frac{\log(\mathbf{N})}{\epsilon^2} \rceil)$ nodes are sampled from the total $\mathbf{N}$ nodes. For each sampled node, we compute its connectivity with all $\mathbf{N}$ nodes to identify approximate connected components. As a result, when $\mathbf{N} > \lceil \frac{\log(\mathbf{N})}{\epsilon^2} \rceil$, the time complexity is:
\begin{equation}
    O(\mathbf{N}'\cdot\mathbf{N}\alpha(\mathbf{N})) = O(\min(\mathbf{N}, \lceil \frac{\log(\mathbf{N})}{\epsilon^2}\rceil)\cdot\mathbf{N}\alpha(\mathbf{N})) = O(\log(\mathbf{N})\cdot\mathbf{N}\alpha(\mathbf{N})).
\end{equation}
Specifically, when the error tolerance $\epsilon$ is set to 0.05, the overall time complexity of the algorithm reduces to $O(\log(\mathbf{N}) \cdot \mathbf{N} \cdot \alpha(\mathbf{N}))$ once the number of tokens exceeds 1200.

\end{document}